\documentclass[11pt,a4paper,onecolumn]{personalarxiv}

\usepackage{amsmath,amsfonts,bm}

\def\eqref#1{equation~\ref{#1}}

\def\1{\bm{1}}

\DeclareMathAlphabet{\mathsfit}{\encodingdefault}{\sfdefault}{m}{sl}
\SetMathAlphabet{\mathsfit}{bold}{\encodingdefault}{\sfdefault}{bx}{n}

\renewcommand{\eqref}[1]{(\ref{#1})}
\usepackage{algorithm}
\usepackage{algorithmic}
\usepackage{subcaption}

\title{Not All Dimensions Are Equal: \ \ \ \ 
Outlier-Aware Attention Linearization with Grouped Quadratic Kernels}
\papershorttitle{Outlier-Aware Attention Linearization}

\author[1]{Weiqi Sun\thanks{These authors contributed equally to this work.}}
\author[2]{Qianyue Wu\protect\footnotemark[1]}
\author[3]{Wenjia Wang\thanks{Corresponding author:
  \href{mailto:wenjiawang@nus.edu.sg}{\texttt{wenjiawang@nus.edu.sg}}}}
\author[1]{Tianyang Hu\thanks{Corresponding author:
  \href{mailto:hutianyang@cuhk.edu.cn}{\texttt{hutianyang@cuhk.edu.cn}}}}
\affil[1]{The Chinese University of Hong Kong, Shenzhen}
\affil[2]{Beijing Normal University, Zhuhai}
\affil[3]{National University of Singapore}
\paperpdfauthor{Weiqi Sun, Qianyue Wu, Wenjia Wang, Tianyang Hu}

\begin{document}
\maketitle
\begin{abstract}
Linearizing softmax attention offers a route to efficient Transformer inference, but preserving pretrained attention remains challenging.
We identify an important source of approximation error overlooked by standard linearization methods: a small number of outlier dimensions contribute disproportionately to query--key dot products and account for a large fraction of the approximation error. 
Motivated by this heterogeneity, we propose \textbf{Outlier-Aware Attention Linearization (OAL)}, a dimension-aware method for constructing quadratic attention approximations.
Starting from a second-order Taylor approximation, we introduce a Hadamard-product formulation that makes individual query--key contributions explicit.
This formulation relaxes globally shared coefficients to support dimension-specific linear terms and cross-dimensional quadratic interactions while preserving separable evaluation. 
To retain this flexibility with fewer parameters, OAL groups dimensions with similar contribution scales to share coefficients, while allowing distinct treatment of the few outliers. 
The resulting positive quadratic kernel is calibrated against pretrained attention outputs and admits linear-time evaluation with respect to sequence length for fixed head dimensions.
On Qwen2.5-0.5B, OAL reduces frozen-attention output
relative MSE by approximately $8.9\%$ compared with DiJiang. 
Replacing middle-layer attention modules and jointly fine-tuning the kernel coefficients with LoRA further lowers WikiText-103 perplexity from $31.71$ to $27.50$ compared with a learned global quadratic, while improving accuracy on PIQA, HellaSwag, and ARC-E. 
These results demonstrate the value of outlier-aware dimension grouping for both attention approximation and pretrained-model conversion. 
The code is available at \href{https://github.com/wuqianyue09/OAL}{GitHub}.
\end{abstract}

\section{Introduction}
\label{sec:introduction}

Softmax attention is central to modern Transformers, but computing all pairwise token interactions incurs quadratic computation as context length grows~\citep{vaswani2017attention}.
Linear attention offers an alternative: separable query and key features permit recurrent evaluation with linear sequence-length complexity for a fixed feature dimension~\citep{katharopoulos2020transformers}.
Approximating softmax attention in pretrained Transformers with linear attention therefore offers a promising route to improving inference efficiency while reusing pretrained representations.
Realizing this potential, however, requires approximations that closely preserve the original attention behavior, particularly when only limited adaptation is available to compensate for approximation errors.

Existing methods approximate the softmax kernel using feature maps that separate query and key computations, or low-order polynomials that admit efficient evaluation~\citep{choromanski2021rethinking,nguyen2026spectraformer,chen2024dijiang,nauen2024taylorshift}.
To better preserve pretrained attention, these approximations can also be learned or adapted to match the original model's attention behavior~\citep{chowdhury2022learning,zhang2025lolcats,lan2025liger}.
Despite these advances, approximation errors are still to be desired under the limited-adaptation setting.
To better understand the remaining approximation error, we examine how it relates to the contributions of individual query--key dimensions.

Our analysis reveals that a small subset of query--key dimensions contributes disproportionately to attention logits (Section \ref{sec:observations}).
Diagnostic experiments further show that approximating these dimensions alone can induce substantial attention-output error.
This suggests that the approximation should have sufficient flexibility to accommodate differences in dimension-wise contributions.
The representative approaches discussed above generally formulate their approximations through vector-level feature mappings or functions of the aggregated query--key score.
Their parameterization, however, is not explicitly organized around the unequal contributions of individual query--key dimensions.
Our findings motivate making these contributions explicit and using their statistics to guide the approximation, so that outlier dimensions can be treated differently from the remaining dimensions.

Therefore, we propose \emph{Outlier-Aware Attention Linearization} (OAL), which explicitly adapts the approximation to dimension-wise heterogeneity in pretrained attention.
We use a quadratic kernel as a tractable starting point for introducing dimension-specific parameters while retaining separable evaluation.
To enable dimension-wise treatment, we introduce a Hadamard-product formulation that makes individual query--key contributions explicit.
This formulation generalizes the globally shared coefficients in a quadratic approximation to dimension-specific linear coefficients and coefficients for cross-dimensional quadratic interactions.
It thereby allows different dimensions to be approximated differently while preserving the separable structure required for linear-time evaluation.

A fully dimension-specific parameterization, however, introduces many coefficients. 
The empirical distribution of dimension-wise contributions suggests that much of this flexibility can be shared.
This observation motivates grouping dimensions with similar contributions and sharing coefficients within groups and across corresponding group pairs.
The resulting parameterization accommodates outlier dimensions while keeping the number of learned coefficients small.
With the groups fixed, we calibrate the resulting positive quadratic kernel against pretrained softmax attention outputs using frozen queries, keys, and values.

On frozen Qwen2.5-0.5B attention states, OAL achieves the lowest error among the evaluated methods, reducing attention-output relative MSE by approximately $8.9\%$ compared with DiJiang~\citep{chen2024dijiang}.
Under a common middle-layer attention replacement protocol with matched LoRA adaptation~\citep{hu2022lora}, OAL outperforms the evaluated converted methods on WikiText-103 perplexity and all three downstream tasks on both Qwen2.5-0.5B and Llama-3.2-1B~\citep{llama2024herd}.
Relative to the learned global quadratic, it lowers Qwen2.5-0.5B perplexity from $31.71$ to $27.50$, although a gap to the LoRA-adapted softmax reference remains on both backbones.
OAL also requires only $O(N)$ attention computation for a sequence of length $N$ and fixed head dimensions, without explicitly constructing the $N\times N$ attention matrix.

Our contributions are threefold:
\begin{itemize}
    \item We characterize dimension-wise heterogeneity in pretrained query--key interactions and its association with attention
    approximation error, motivating contribution-aware kernel design.

    \item 
    We develop a dimension-aware quadratic formulation with separable evaluation, and introduce contribution-based grouping to retain this flexibility with a compact parameterization.

    \item We demonstrate improved frozen-attention fidelity and adapted
    model quality under a matched conversion protocol, and examine the
    role of group-aware flexibility through a diagnostic budget study.
\end{itemize}

\section{Preliminaries}
\label{sec:preliminaries}

\paragraph{Softmax Attention}
For a single head, let $Q,K\in\mathbb{R}^{N\times d_h}$ and
$V\in\mathbb{R}^{N\times d_v}$ denote the query, key, and value matrices,
with sequence length $N$, query--key dimension $d_h$, and value dimension $d_v$.
We write $q_i,k_i,v_i$ for their respective token vectors at position $i$.
Scaled dot-product attention~\citep{vaswani2017attention} is
\[
    \operatorname{Attn}(Q,K,V)
    =
    \operatorname{Softmax}
    \left(
        \frac{QK^\top}{\sqrt{d_h}}
    \right)V,
\]
with row-wise softmax.
Direct evaluation costs $O(N^2(d_h+d_v))$ operations;
explicit storage of the score matrix requires $O(N^2)$ memory.

\paragraph{Linear Attention}
A common class of linear attention methods replaces the pairwise
attention kernel with a separable feature representation
\citep{katharopoulos2020transformers,choromanski2021rethinking}.
Specifically, the softmax kernel can be approximated as
$\exp(q^\top k/\sqrt{d_h})\approx\kappa(q,k)=\phi(q)^\top\phi(k)$,
where $\phi(\cdot) \in \mathbb{R}^{m}$ denotes a feature map.
Substituting this factorization into normalized causal attention yields
\begin{equation}
    \widehat{y}_i
    =
    \frac{
        \left(
            \sum_{j \leq i} v_j \phi(k_j)^\top
        \right)
        \phi(q_i)
    }{
        \phi(q_i)^\top
        \left(
            \sum_{j \leq i} \phi(k_j)
        \right)
    }.
    \label{eq:linear_attention}
\end{equation}
Here, $\widehat y_i$ denotes the approximate attention output at position $i$.
The key--value and key statistics in parentheses can be accumulated
incrementally over the causal prefix.
This reassociation avoids explicitly constructing the $N \times N$ attention matrix and gives linear scaling with sequence length for a fixed feature dimension and memory size~\citep{gu2023mamba, sun2023retentive, peng2023rwkv, chen2024exact}. 
Different linearization methods therefore mainly differ in the choice
or construction of the feature map $\phi(\cdot)$ and the resulting
approximation to the softmax kernel.

\paragraph{Softmax Kernel Approximation}
Separable softmax approximations can be obtained by constructing
feature maps directly
\citep{katharopoulos2020transformers,
choromanski2021rethinking,
peng2021random}
or by expanding a low-order polynomial in the attention score into
separable terms. We focus on the latter. Polynomial kernels based on
low-order Taylor approximations have been explored in efficient
attention and language modeling
\citep{arora2024based,nauen2024taylorshift}.
For a scalar attention score $s$, the second-order Taylor approximation is
$\exp(s)\approx 1+s+\tfrac12 s^2$.
Allowing its coefficients to vary gives the global quadratic
$p(s)=a+bs+cs^2$,
where $a$, $b$, and $c$ control the constant, linear, and quadratic
terms. This is the starting point for our dimension-aware kernel.

\section{Dimension-Wise Heterogeneity in Pretrained Attention}
\label{sec:observations}

Motivated by activation outliers in pretrained
Transformers~\citep{kovaleva2021bert,dettmers2022llmint8,massive_activation, li2026transformers, li2026structural},
where a small number of
activation or representation dimensions can attain unusually large
magnitudes or exert disproportionate influence on model behavior,
we examine whether query--key contributions are similarly concentrated
in a few dimensions and whether this concentration is associated with
attention-approximation error. 

\begin{figure}[t]
    \centering
    \begin{subfigure}[t]{0.32\linewidth}
        \centering
        \includegraphics[width=\linewidth]{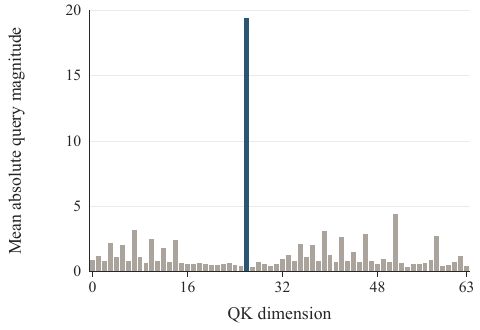}
        \caption{Query magnitude}
        \label{fig:qwen_query_magnitude}
    \end{subfigure}
    \hfill
    \begin{subfigure}[t]{0.32\linewidth}
        \centering
        \includegraphics[width=\linewidth]{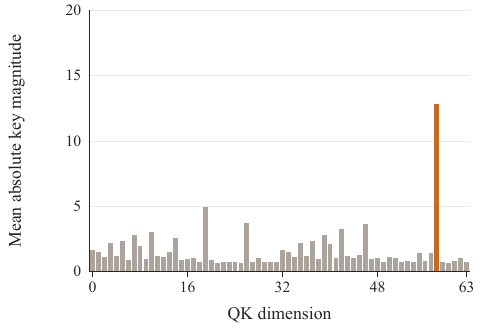}
        \caption{Key magnitude}
        \label{fig:qwen_key_magnitude}
    \end{subfigure}
    \hfill
    \begin{subfigure}[t]{0.32\linewidth}
        \centering
        \includegraphics[width=\linewidth]{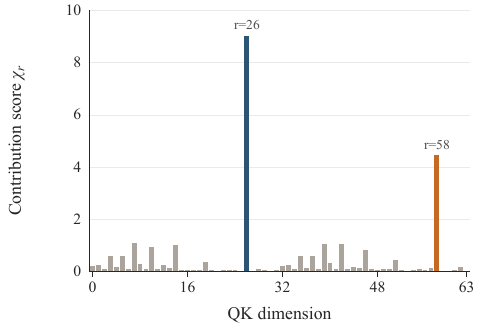}
        \caption{Dimension-wise contribution}
        \label{fig:qwen_qk_contribution}
    \end{subfigure}
    \caption{
    \textbf{Dimension-wise heterogeneity in a representative Qwen2.5-0.5B attention head.}
    (a) Mean absolute query magnitude across QK dimensions.
    (b) Mean absolute key magnitude across the corresponding KV head.
    (c) Contribution score $\chi_r$ defined in Eq.~\eqref{eq:qk_contribution}.
    Dimensions $r=26$ and $r=58$ are highlighted across panels.
    }
    \label{fig:qk_outliers}
\end{figure}

\subsection{Query--Key Outliers and Dimension-Wise Contributions}
\label{sec:qk_contribution}

Figure~\ref{fig:qk_outliers} shows the post-RoPE query and key magnitudes
for a representative Qwen2.5-0.5B head. Since either magnitude alone
does not directly measure a dimension's contribution to attention logits, we
analyze their coordinate-wise products for each layer--head pair,
suppressing layer and head indices below.
For calibration input $p$, let $N_p$ denote its sequence length, and let
$q_{p,i},k_{p,j}\in\mathbb{R}^{d_h}$ denote the post-RoPE query and key
vectors at positions $i$ and $j$. Their $r$-th coordinates are denoted
by $q_{p,i,r}$ and $k_{p,j,r}$. We define the causal pair set as
$\Omega_p=\{(i,j):1\leq j\leq i\leq N_p\}$.

We define the contribution of dimension $r$ on input $p$ and its
equally weighted average over the $P$ calibration inputs as
\begin{equation}
    \chi_{p,r}
    =
    \frac{1}{|\Omega_p|}
    \sum_{(i,j)\in\Omega_p}
    \left|
        \frac{q_{p,i,r}k_{p,j,r}}{\sqrt{d_h}}
    \right|,
    \qquad
    \chi_r
    =
    \frac{1}{P}
    \sum_{p=1}^{P} \chi_{p,r}.
    \label{eq:qk_contribution}
\end{equation}

The contribution profile in Figure~\ref{fig:qk_outliers} is concentrated
on a few dimensions. The highlighted dimensions, $r=26$ and $r=58$,
have high contribution scores and correspond to prominent coordinates
in the query or key representations. A similarly
concentrated dimension-wise contribution profile is observed in
GPT-2 Medium; additional cross-model evidence is provided in
Appendix~\ref{app:qk_heterogeneity}.

\subsection{Approximation Error Concentration on High-Contribution Dimensions}
\label{sec:contribution_error}

We next test whether approximating only the high-contribution
dimensions retains a substantial fraction of the full-head output error.
For each evaluated attention head, we select a small subset
$\mathcal S$ of high-contribution dimensions according to the
pretrained contribution scores. We apply each approximation only to
$\mathcal S$ while keeping the remaining dimensions exact. Let
$\mathcal E_{\mathcal S}$ denote the resulting attention-output MSE and
$\mathcal E_{\mathrm{full}}$ the error when the same method is applied
to the full head. We report the retained-error ratio
$R_{\mathcal S}=\mathcal E_{\mathcal S}/\mathcal E_{\mathrm{full}}$.
For the Gaussian controls, pretrained Q/K representations are replaced
with i.i.d.\ $\mathcal N(0,1)$ samples and the same contribution-based
grouping and intervention procedures are applied. Further metric and
control details are provided in
Appendix~\ref{app:qk_heterogeneity}.

\begin{table}[t]
    \centering
    \normalsize
    \caption{
    \textbf{Approximation error concentration on high-contribution
    dimensions.}
    The selected subsets account for approximately $5.5\%$ of the
    evaluated head dimensions in aggregate. We report the retained-error
    ratio $R_{\mathcal S}$ (\%) for the high-contribution subsets and for
    Gaussian controls obtained by replacing pretrained Q/K with
    i.i.d.\ $\mathcal N(0,1)$ samples while keeping the remaining protocol
    unchanged. Higher values indicate stronger error concentration.
    }
    \label{tab:error_concentration}
    \setlength{\tabcolsep}{5pt}
    \begin{tabular}{lccc}
        \toprule
        Method
        & \shortstack{High-contrib. error (\%)}
        & \shortstack{Gaussian control (\%)} \\
        \midrule
        Performer
        & 37.8 & 4.7\\

        SpectraFormer
        & 58.1 & 18.1\\

        DiJiang
        & 64.8 & 19.4\\

        TaylorShift
        & 63.9 & 13.6\\
        \bottomrule
    \end{tabular}
\end{table}

Approximating only about $5.5\%$ of the dimensions retains a substantial
fraction of the full-head attention-output MSE across all evaluated
methods, with retained-error ratios consistently above the corresponding
Gaussian controls (Table~\ref{tab:error_concentration}). In a separate random-coordinate control with 1,000 cardinality-matched
subsets sampled from the pretrained bulk dimensions, none of the null
replicates exceeded the observed retained-error ratio for any method,
yielding empirical $p<0.001$ in all cases. These retained-error
ratios suggest an association between high contribution scores and
attention-output approximation error.
Appendix~\ref{app:qk_heterogeneity}
provides additional control analyses and stability analyses across held-out
inputs and sequence lengths for the representative Qwen2.5-0.5B head.
These observations motivate the contribution-aware parameter sharing
in Section~\ref{sec:method}, which allows dimensions with different
contribution scales to use different approximation coefficients.

\section{Outlier-Aware Attention Linearization}
\label{sec:method}

We introduce dimension-specific linear and cross-dimensional quadratic
coefficients, with a Hadamard--diagonal (HD) formulation enabling exact
kernel evaluation in linear time with respect to sequence length for
fixed head dimensions. Contribution-based grouping then shares
coefficients across dimensions and dimension pairs; enforcing positive
kernel weights yields OAL.

\subsection{Dimension-Aware Linearization}
\label{sec:dimension_aware_linearization}

For a query--key pair $q,k\in\mathbb R^{d_h}$, let
$x_r=q_rk_r/\sqrt{d_h}$ denote the contribution of dimension $r$, so
that the attention score is $s=\sum_r x_r$. A global quadratic
approximation to $e^s$ can be written as
\[
    p(s)=a+bs+cs^2
    =a+\sum_{r=1}^{d_h}bx_r
      +\sum_{r=1}^{d_h}\sum_{t=1}^{d_h}cx_rx_t.
\]
All linear contributions therefore share $b$, and all quadratic
interactions share $c$.
TaylorShift uses this scalar polynomial structure with coefficients
prescribed by the Taylor expansion \citep{nauen2024taylorshift}.
To accommodate the dimension-wise heterogeneity observed in
Section~\ref{sec:observations}, we relax this sharing and allow separate
coefficients for each dimension and dimension pair:
\begin{equation}
    \kappa_{\mathrm{dim}}(q,k)
    =a+\sum_{r=1}^{d_h} b_rx_r
      +\sum_{r=1}^{d_h}\sum_{t=1}^{d_h}C_{rt}x_rx_t,
    \qquad C=C^\top.
    \label{eq:dimension_aware_quadratic}
\end{equation}
Here, $r$ and $t$ index the $d_h$ query--key dimensions;
$b_r$ weights the contribution of dimension $r$, $C_{rr}$ weights
its squared term, and the off-diagonal entries of $C$ weight
cross-dimensional interactions. Setting $b_r=b$ and $C_{rt}=c$ recovers
the global quadratic. Degree two is the lowest degree that retains both
squared terms and cross-dimensional interactions, using $O(d_h^2)$
features.

\paragraph{Hadamard-diagonal formulation}
Evaluating the quadratic kernel explicitly over all query--key pairs
still incurs quadratic cost in sequence length. To avoid this pairwise
computation, we use the following Hadamard--diagonal (HD) identity for
the shared-coefficient quadratic term. For value channel $V_{:,\ell}$,
let $D_{\ell}\in\mathbb R^{N\times N}$ be the diagonal matrix with entries
$(D_{\ell})_{jj}=V_{j,\ell}$. Then
\begin{equation}
    \big[(QK^\top)\odot(QK^\top)\big]V_{:,\ell}
    =\operatorname{diag}\!\left(
      Q\big[K^\top D_{\ell}K\big]Q^\top
    \right).
    \label{eq:hd_diagonal_identity}
\end{equation}
Here, $\odot$ denotes the Hadamard product, and $\operatorname{diag}$
extracts the diagonal entries of a matrix as a vector. By first
accumulating $K^\top D_{\ell}K$ and then evaluating only the diagonal entries,
the quadratic term can be computed in $O(Nd_h^2)$ time per value channel
without forming an $N\times N$ matrix. For fixed head dimensions, this
cost is linear in sequence length.

The $(r,t)$ entry of $K^\top D_{\ell}K$ is
$\sum_jV_{j,\ell}k_{j,r}k_{j,t}$, which sums value-weighted key products
over token positions $j$. For a query at position $i$, restricting this
sum to $j\leq i$ gives causal prefix statistics. The attention denominator
uses the same kernel weights as the numerator, but replaces each value
with $1$. Thus, augmenting the values as $u_j=[v_j;1]$ allows both to be
accumulated together. Including the constant and linear kernel terms
gives the states
\begin{equation}
    H_{0,i}=\sum_{j\leq i}u_j,\qquad
    H_{1,i,r}=\sum_{j\leq i}k_{j,r}u_j,\qquad
    H_{2,i,rt}=\sum_{j\leq i}k_{j,r}k_{j,t}u_j.
    \label{eq:oal_hd_states}
\end{equation}
Each $H_{2,i,rt}$ is a $(d_v+1)$-dimensional vector.

The query combines these statistics using the coefficients $a$, $b_r$,
and $C_{rt}$ from Eq.~\eqref{eq:dimension_aware_quadratic}:
\begin{equation}
    \begin{aligned}
        \widetilde u_i
        &=aH_{0,i}+\frac{1}{\sqrt{d_h}}\sum_r b_rq_{i,r}H_{1,i,r}
          +\frac{1}{d_h}\sum_{r,t}C_{rt}q_{i,r}q_{i,t}H_{2,i,rt},\\
        \widehat y_i&=\widetilde u_{i,1:d_v}/\widetilde u_{i,d_v+1}.
    \end{aligned}
    \label{eq:oal_hd_readout}
\end{equation}
This recurrence evaluates the specified quadratic kernel exactly
without forming a token-by-token matrix
(Appendix~\ref{app:oal_realization}). The last channel supplies the
normalization denominator, whose positivity follows from the OAL
parameterization in Section~\ref{sec:oal_algorithm}.

A fully dimension-specific quadratic has $O(d_h^2)$ coefficients.
Contribution-based sharing reduces this parameter count
while retaining the same states and readout.

\subsection{Outlier-Aware Dimension Grouping}
\label{sec:dimension_grouping}

\begin{figure}[t]
    \centering
    \begin{subfigure}[t]{0.49\linewidth}
        \centering
        \includegraphics[width=\linewidth]
        {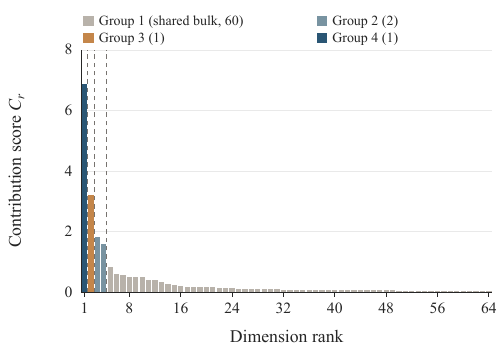}
        \caption{Representative group assignment.}
        \label{fig:representative_grouping}
    \end{subfigure}
    \hfill
    \begin{subfigure}[t]{0.49\linewidth}
        \centering
        \includegraphics[width=\linewidth]
        {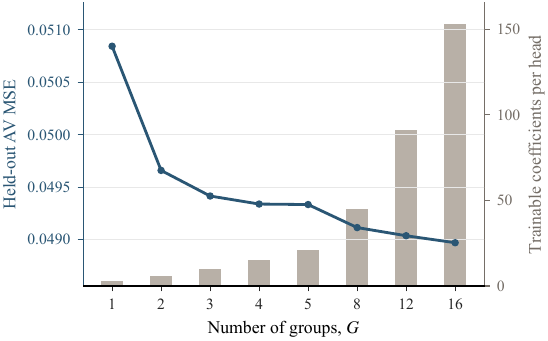}
        \caption{Group-budget trade-off.}
        \label{fig:group_count_sensitivity}
    \end{subfigure}

    \caption{
    \textbf{Contribution-aware dimension grouping.}
    (a) Sorted contribution scores $\chi_r$ and the resulting OAL groups for
    the representative Qwen2.5-0.5B Layer 4, Head 0.
    (b) Attention-output MSE and trainable coefficients per head under
    fixed group budgets for the same head. All budgets use the same data
    splits and coefficient-fitting protocol.
    }
    \label{fig:grouping_overview}
\end{figure}

The contribution profile in Figure~\ref{fig:qk_outliers} suggests a
compact sharing structure: low-contribution dimensions can share
parameters, while the high-contribution tail may benefit from finer
parameterization. Figure~\ref{fig:grouping_overview}(a) illustrates a
partition for the representative head in Section~\ref{sec:qk_contribution}.
For a partition $\{\mathcal G_g\}_{g=1}^{G}$ of the dimensions, recall that $x_r=q_rk_r/\sqrt{d_h}$ as the contribution of dimension $r$ and
$s_g=\sum_{r\in\mathcal G_g}x_r$ as the total contribution of group $g$.
Thus, $s=\sum_{g=1}^{G}s_g$. Sharing $b_r$ within each group and
$C_{rt}$ within each group pair reduces the number of learned
coefficients to $O(G^2)$ while preserving the HD evaluation.

For each layer and query head, we sort dimensions by their calibration
contribution scores $\chi_r$ and identify large gaps in the high-contribution
tail that remain stable under prompt-level bootstrap resampling,
following resampling-based stability assessment
\citep{suzuki2006pvclust,hennig2007clusterwise}. These boundaries determine
the group count, with at least two groups. Groups remain
fixed during coefficient fitting and model adaptation. The procedure
yields $2$--$5$ groups per head for Qwen2.5-0.5B and $2$--$4$ for
Llama-3.2-1B; full details are provided in
Appendix~\ref{app:grouping_calibration_details}.

As a separate diagnostic, Figure~\ref{fig:grouping_overview}(b) illustrates
the trade-off between attention-output MSE and parameter count across
fixed group budgets for the representative head. For this head, a small
number of groups captures most of the error reduction, with diminishing
gains as the budget increases.

\subsection{The OAL Algorithm}
\label{sec:oal_algorithm}

OAL uses a positive quadratic kernel whose coefficients are shared
according to contribution-based dimension groups. We specify this
parameterization below and combine it with the HD formulation to
summarize the complete method in Algorithm~\ref{alg:oal}.

\paragraph{Positive group-aware kernel}
Let $g(r)$ denote the group containing dimension $r$.
To enforce positive kernel weights while sharing coefficients within
groups and group pairs, we introduce a learnable lower-triangular matrix
$L\in\mathbb R^{(G+1)\times(G+1)}$ for each layer and query head and
set $M=LL^\top$.
The coefficients are $a=\epsilon+M_{00}$, $b_r=2M_{0,g(r)}$,
and $C_{rt}=M_{g(r),g(t)}$, with $\epsilon>0$.
With $\bar{\mathbf s}=[1,s_1,\ldots,s_G]^\top$, this parameterization gives
\begin{equation}
    \kappa_{\mathrm{OAL}}(q,k)
    =\epsilon+\bar{\mathbf s}^{\top}LL^{\top}\bar{\mathbf s},
    \qquad \epsilon>0,
    \label{eq:oal_kernel}
\end{equation}
Since $LL^\top\succeq0$, we have
$\kappa_{\mathrm{OAL}}(q,k)\geq\epsilon>0$, ensuring a positive
normalization denominator.

With the groups fixed, we learn $L$ by minimizing the squared error
between normalized OAL and softmax attention outputs on an offline
calibration set, using frozen pretrained queries, keys, and values.
Calibration details are provided in
Appendix~\ref{app:grouping_calibration_details}. The calibrated factors
initialize subsequent joint adaptation with LoRA, during which the
groups remain fixed.

\paragraph{Causal evaluation}
The coefficient mapping above expresses the grouped kernel in the
dimension-wise form of Eq.~\eqref{eq:dimension_aware_quadratic}, so it
can be evaluated using the HD states and readout in
Eqs.~\eqref{eq:oal_hd_states} and~\eqref{eq:oal_hd_readout}.
This computation is exact for the specified quadratic kernel;
approximation arises only from replacing the softmax kernel.
Algorithm~\ref{alg:oal} summarizes the offline calibration and causal
evaluation stages. The equivalence proof is provided in
Appendix~\ref{app:oal_realization}.

\begin{algorithm}[t]
    \caption{Outlier-Aware Attention Linearization (OAL)}
    \label{alg:oal}
    \begin{algorithmic}[1]
        \REQUIRE Calibration data and pretrained attention states; evaluation inputs $Q,K,V$; $\epsilon>0$
        \STATE \textbf{Offline calibration (per layer and query head)}
        \STATE Determine group map $g$ from contribution statistics (Section~\ref{sec:dimension_grouping})
        \STATE Fit $L$ by output matching with $g$ and pretrained states fixed (Eq.~\eqref{eq:oal_calibration})
        \STATE \textbf{Causal evaluation (fixed $g,L$)}
        \STATE Compute $a,b_r,C_{rt}$ from $L,g,\epsilon$; initialize $H_0,H_1,H_2=0$
        \FOR{$i=1,\ldots,N$}
            \STATE $u_i\leftarrow[v_i;1]$; $H_0\leftarrow H_0+u_i$
            \STATE $H_{1,r}\leftarrow H_{1,r}+k_{i,r}u_i$ for all $r$
            \STATE $H_{2,rt}\leftarrow H_{2,rt}+k_{i,r}k_{i,t}u_i$ for all $r,t$
            \STATE Compute $\widehat y_i$ using Eq.~\eqref{eq:oal_hd_readout}
        \ENDFOR
        \RETURN $\widehat Y=[\widehat y_1,\ldots,\widehat y_N]^\top$
    \end{algorithmic}
\end{algorithm}

\subsection{Complexity Analysis}
\label{sec:oal_complexity}

Direct pairwise evaluation costs
$O(N^2(d_h+G^2+d_v))$, whereas the causal evaluation stage of
Algorithm~\ref{alg:oal} costs
$O(Nd_h^2(d_v+1))$.
The HD states require $O(d_h^2(d_v+1))$ space, excluding inputs and
outputs. For fixed head dimensions, computation is linear in sequence
length and state space is independent of it. Grouping reduces the learned
parameter count to $O(G^2)$, while the implementation retains
dimension-pair states. Detailed counts appear in
Appendix~\ref{app:oal_realization}.

Figure~\ref{fig:oal_runtime_memory} reports causal attention forward
latency and peak GPU memory under the setup in
Appendix~\ref{app:runtime_benchmark}. OAL exhibits approximately linear
runtime growth at longer sequences. With the same HD block-GEMM
backend, OAL ($G=5$) achieves latency and peak memory comparable to
the single-group Learned Global baseline ($G=1$), suggesting that the
additional coefficient flexibility introduces little operator overhead
in this configuration. Cross-method comparisons reflect the evaluated
implementations, which differ in numerical precision and optimization level.
FlashAttention remains faster and more memory-efficient throughout
the tested range.

\begin{figure}[t]
    \centering
    \includegraphics[width=\linewidth]{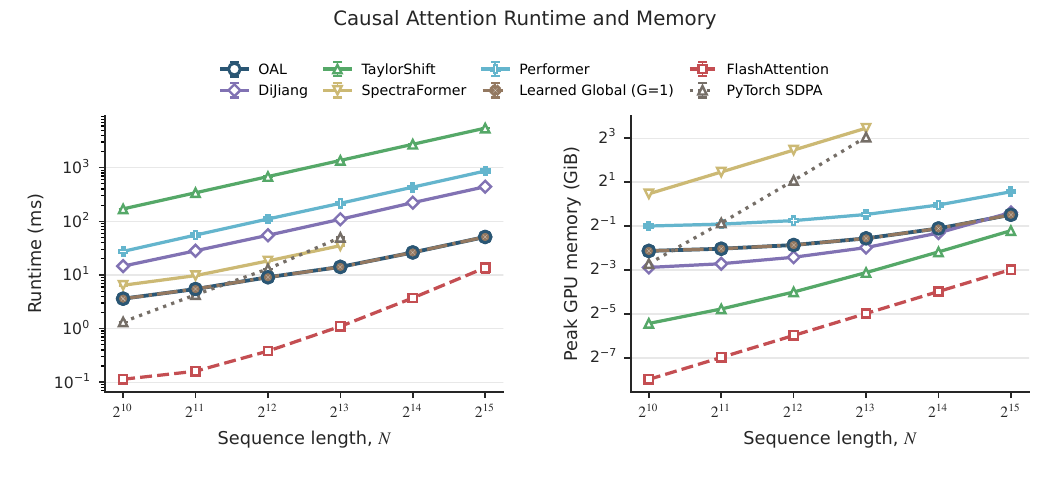}
    \caption{\textbf{Causal attention forward runtime and memory.}
    Forward-only measurements use an A100-SXM4-40GB with BF16 inputs.
    OAL ($G=5$) and Learned Global ($G=1$) share
    the HD block-GEMM backend.
    Memory denotes total peak PyTorch-allocated tensor memory.
    SpectraFormer and SDPA (math backend) are capped at $N=8192$
    by a pre-run memory guard. Implementation and measurement details
    appear in Appendix~\ref{app:runtime_benchmark}.}
    \label{fig:oal_runtime_memory}
\end{figure}

\section{Experiments}
\label{sec:experiments}

We first evaluate OAL's approximation fidelity on frozen pretrained
attention states. We then examine whether OAL supports better
language-modeling and downstream performance after attention replacement
under a common, limited LoRA adaptation budget.

\subsection{Frozen-Attention Approximation Fidelity}
\label{sec:frozen_fidelity}

To isolate approximation error before model adaptation, we compare OAL
and existing attention linearization methods on identical frozen queries,
keys, and values from Qwen2.5-0.5B Base, with queries and keys taken after
RoPE. The evaluation covers all query heads in Layers 4, 12, and 20
at sequence length $N=4096$.

Approximation error is measured by MSE and relative MSE (RelMSE)
against the corresponding softmax outputs, both before and after the
output projection $W_o$. Exact metric definitions
and aggregation procedures are provided in
Appendix~\ref{app:frozen_metrics}.

\begin{table}[t]
    \centering
    \normalsize
    \caption{
    \textbf{Frozen-attention approximation fidelity on Qwen2.5-0.5B.}
    Results cover all query heads in Layers 4, 12, and 20 at sequence
    length $4096$. Lower is better for all metrics.
    }
    \label{tab:frozen_fidelity}
    \setlength{\tabcolsep}{4pt}
    \begin{tabular}{lcccc}
        \toprule
        Method
        & \shortstack{Before $W_o$\\MSE $\downarrow$}
        & \shortstack{Before $W_o$\\RelMSE $\downarrow$}
        & \shortstack{After $W_o$\\MSE $\downarrow$}
        & \shortstack{After $W_o$\\RelMSE $\downarrow$} \\
        \midrule
        Performer
        & 0.103 & 0.569 & 0.038 & 0.710 \\

        SpectraFormer
        & 0.106 & 0.585 & 0.041 & 0.772 \\

        DiJiang 
        & 0.086 & 0.474 & 0.033 & 0.631 \\

        TaylorShift
        & 0.088 & 0.490 & 0.034 & 0.649 \\

        \midrule
        OAL
        & \textbf{0.078}
        & \textbf{0.432}
        & \textbf{0.031}
        & \textbf{0.595} \\
        \bottomrule
    \end{tabular}
\end{table}

OAL achieves the lowest MSE and RelMSE both before and after the output
projection $W_o$ (Table~\ref{tab:frozen_fidelity}). DiJiang is the
strongest evaluated baseline under all four metrics.

\subsection{End-to-End Conversion and Downstream Performance}
\label{sec:end_to_end}

The conversion experiments use Qwen2.5-0.5B Base and
Llama-3.2-1B Base. We compare OAL with a softmax reference, a learned
global quadratic, Performer, SpectraFormer, DiJiang, and TaylorShift.
The learned global quadratic is the single-group ($G=1$) instance of
OAL and shares its calibration and adaptation protocol, providing a
controlled comparison of global and contribution-based coefficient
sharing. Following the hybrid conversion setting explored in
LoLCATs~\citep{zhang2025lolcats}, we retain softmax attention in
the initial and final layers. Our layer-wise activation analysis
motivates converting the intervening layers, where large activations
persist across depth (Appendix~\ref{app:conversion_setup}). All converted
methods use the same replacement range within each backbone.

All methods, including the softmax reference, use the same
LoRA configurations, training data, and update budgets within each
backbone. Adaptation uses
next-token prediction on
WikiText-103. For OAL, the calibrated kernel coefficients are jointly
optimized with the LoRA parameters, while the dimension groups remain
fixed. We adapt Performer, SpectraFormer, DiJiang, and TaylorShift from
their official source code; modifications are detailed in
Appendix~\ref{app:baseline_adaptations}, and shared conversion and
adaptation settings in Appendix~\ref{app:conversion_setup}.
Table~\ref{tab:end_to_end_results} reports language-modeling and
downstream results for both backbones. Evaluation protocols are detailed
in Appendix~\ref{app:evaluation_protocol}.

\begin{table}[t]
    \centering
    \normalsize
    \setlength{\tabcolsep}{3pt}
    \caption{Model-level quality after LoRA adaptation on two backbones.
    We report WikiText-103 test PPL (lower is better) and zero-shot
    normalized accuracy (\%, higher is better).
    Bold highlights the leading results among converted methods
    within each backbone.}
    \label{tab:end_to_end_results}
    \begin{tabular}{lcccc@{\hspace{10pt}}cccc}
        \toprule
        & \multicolumn{4}{c}{Qwen2.5-0.5B}
        & \multicolumn{4}{c}{Llama-3.2-1B} \\
        \cmidrule(lr){2-5} \cmidrule(lr){6-9}
        Method & PPL & PIQA
        & HellaSwag & ARC-E
        & PPL & PIQA
        & HellaSwag & ARC-E \\
        \midrule
        Softmax (LoRA) & 10.89 & 70.18 & 51.38 & 62.63
                       & 8.19 & 74.92 & 62.84 & 59.43 \\
        \midrule
        Performer     & 37.03 & 57.78 & 30.37 & 38.93
                      & 50.39 & 57.07 & 27.42 & 32.58 \\
        SpectraFormer & 34.78 & 59.19 & 31.77 & 41.84
                      & 45.09 & 59.36 & 30.41 & 36.49 \\
        DiJiang       & 36.05 & 59.90 & 31.47 & 41.41
                      & 56.27 & 58.92 & 28.75 & 36.32 \\
        TaylorShift   & 39.50 & 59.90 & 31.18 & 40.78
                      & 62.67 & 57.40 & 28.95 & 35.77 \\
        Learned global & 31.71 & 61.15 & 32.43 & 41.41
                       & 36.03 & 57.40 & 35.02 & 33.46 \\
        OAL           & \textbf{27.50} & \textbf{64.69}
                      & \textbf{37.20} & \textbf{50.38}
                      & \textbf{32.99} & \textbf{63.66}
                      & \textbf{35.42} & \textbf{41.67} \\
        \bottomrule
    \end{tabular}
\end{table}

OAL achieves the lowest WikiText-103 perplexity and highest normalized
accuracy on all three downstream tasks among the evaluated converted
models on both backbones (Table~\ref{tab:end_to_end_results}).
The LoRA-adapted softmax reference remains stronger on every metric,
indicating a substantial conversion gap.
Relative to the learned global quadratic ($G=1$), OAL reduces perplexity
from $31.71$ to $27.50$ on Qwen2.5-0.5B and from $36.03$ to $32.99$
on Llama-3.2-1B. These gains support the effectiveness of contribution-based
grouping over globally shared coefficients under the evaluated protocol.

\section{Related Work}
\label{sec:related_work}

\paragraph{Kernel Linearization and Pretrained-Model Conversion}

Kernel methods such as Performer, SpectraFormer, and DiJiang approximate
softmax attention with feature representations that permit linear-time
evaluation \citep{choromanski2021rethinking,nguyen2026spectraformer,chen2024dijiang}.
T2R converts pretrained Transformers through attention replacement and
finetuning \citep{kasai2021finetuning}.
Hedgehog learns MLP feature maps by matching softmax attention weights
\citep{zhang2024hedgehog}, while LoLCATs combines attention transfer
with low-rank adaptation \citep{zhang2025lolcats}.
OAL shares the goal of preserving pretrained behavior, but organizes
kernel coefficients by dimension-wise contribution, providing a structured
alternative to learned MLP feature maps. Our comparison scope is detailed
in Appendix~\ref{app:baseline_adaptations}.

\paragraph{Polynomial Attention}

In language models trained from scratch, Based uses a second-order
Taylor kernel \citep{arora2024based}, while ReBased modifies the quadratic
kernel with query/key normalization and learnable affine transformations
\citep{aksenov2024rebased}.
PolySketchFormer approximates higher-degree polynomial attention through
sketching \citep{kacham2024polysketchformer}.
OAL targets pretrained softmax recovery under limited adaptation.
It retains original coordinate products and learns
coefficients for linear terms and quadratic interactions, with
sharing determined by contribution statistics.

\paragraph{Activation and Attention Outliers}

Massive activations and attention sinks reveal different forms of
nonuniformity in pretrained Transformers: unusually large hidden
activations and persistent attention to particular positions,
respectively \citep{massive_activation,attention_sink, li2026structural, li2026transformers}.
Systematic Outliers examines how alignment in specific query
and key dimensions produces large attention weights
\citep{an2025systematic}.
Building on these observations, we examine how concentrated query--key
contributions relate to attention-linearization error, using this
association to guide coefficient sharing.

\section{Discussion}
\label{sec:discussion}

We introduce Outlier-Aware Attention Linearization (OAL), a
contribution-aware method for linearizing pretrained softmax attention.
Our analysis shows that approximation error is concentrated on a small
subset of high-contribution query--key dimensions, motivating non-uniform
parameter sharing. OAL captures this structure with a compact grouped
quadratic kernel and supports exact recurrent evaluation with linear
sequence-length complexity. It improves frozen-attention fidelity on
Qwen2.5-0.5B and model-level performance after LoRA adaptation on both
Qwen2.5-0.5B and Llama-3.2-1B. Our experiments are limited to 0.5B--1B
backbones, so performance at larger scales remains unverified. Future
work will evaluate OAL on larger and longer-context models and explore
more compact implementations of the quadratic recurrent state.

\section*{AI Use Statement}

We used generative AI tools to aid and polish writing, including
language editing and improving the clarity and organization of the
manuscript. We also used generative AI tools for retrieval and
discovery, such as finding related work and assisting with literature
search. All AI-assisted content was reviewed and verified by the
authors, who take full responsibility for the final manuscript.

\bibliography{references}

@inproceedings{vaswani2017attention,
  author = {Vaswani, Ashish and Shazeer, Noam and Parmar, Niki and Uszkoreit, Jakob and Jones, Llion and Gomez, Aidan N. and Kaiser, {\L}ukasz and Polosukhin, Illia},
  title = {Attention Is All You Need},
  booktitle = {Advances in Neural Information Processing Systems},
  volume = {30},
  year = {2017},
  url = {https://proceedings.neurips.cc/paper/2017/hash/3f5ee243547dee91fbd053c1c4a845aa-Abstract.html}
}

@article{li2026structural,
  title={The Structural Origin of Attention Sink: Variance Discrepancy, Super Neurons, and Dimension Disparity},
  author={Li, Siquan and Jiang, Kaiqi and Sun, Jiacheng and Hu, Tianyang},
  journal={arXiv preprint arXiv:2605.06611},
  year={2026}
}

@inproceedings{peng2023rwkv,
  title={Rwkv: Reinventing rnns for the transformer era},
  author={Peng, Bo and Alcaide, Eric and Anthony, Quentin and Albalak, Alon and Arcadinho, Samuel and Biderman, Stella and Cao, Huanqi and Cheng, Xin and Chung, Michael and Derczynski, Leon and others},
  booktitle={Findings of the association for computational linguistics: EMNLP 2023},
  pages={14048--14077},
  year={2023}
}

@article{sun2023retentive,
  title={Retentive network: A successor to transformer for large language models},
  author={Sun, Yutao and Dong, Li and Huang, Shaohan and Ma, Shuming and Xia, Yuqing and Xue, Jilong and Wang, Jianyong and Wei, Furu},
  journal={arXiv preprint arXiv:2307.08621},
  year={2023}
}

@article{gu2023mamba,
  title={Mamba: Linear-time sequence modeling with selective state spaces},
  author={Gu, Albert and Dao, Tri},
  journal={arXiv preprint arXiv:2312.00752},
  year={2023}
}

@article{chen2024exact,
  title={Exact conversion of in-context learning to model weights in linearized-attention transformers},
  author={Chen, Brian K and Hu, Tianyang and Jin, Hui and Lee, Hwee Kuan and Kawaguchi, Kenji},
  journal={arXiv preprint arXiv:2406.02847},
  year={2024}
}

@article{li2026transformers,
  title={Transformers are born biased: Structural inductive biases at random initialization and their practical consequences},
  author={Li, Siquan and Tong, Yao and Wang, Haonan and Hu, Tianyang},
  journal={arXiv preprint arXiv:2602.05927},
  year={2026}
}

@inproceedings{katharopoulos2020transformers,
  author = {Katharopoulos, Angelos and Vyas, Apoorv and Pappas, Nikolaos and Fleuret, Fran{\c{c}}ois},
  title = {Transformers are {RNN}s: Fast Autoregressive Transformers with Linear Attention},
  booktitle = {Proceedings of the 37th International Conference on Machine Learning},
  series = {Proceedings of Machine Learning Research},
  volume = {119},
  pages = {5156--5165},
  publisher = {PMLR},
  year = {2020},
  url = {https://proceedings.mlr.press/v119/katharopoulos20a.html}
}

@inproceedings{zhang2024hedgehog,
  author = {Zhang, Michael and Bhatia, Kush and Kumbong, Hermann and R{\'e}, Christopher},
  title = {The Hedgehog \& the Porcupine: Expressive Linear Attentions with Softmax Mimicry},
  booktitle = {International Conference on Learning Representations},
  year = {2024},
  url = {https://openreview.net/forum?id=4g02l2N2Nx}
}

@inproceedings{zhang2025lolcats,
  author = {Zhang, Michael and Arora, Simran and Chalamala, Rahul and Spector, Benjamin and Wu, Alan and Ramesh, Krithik and Singhal, Aaryan and Re, Christopher},
  title = {{LoLCATs}: On Low-Rank Linearizing of Large Language Models},
  booktitle = {International Conference on Learning Representations},
  year = {2025},
  pages = {46200--46253},
  url = {https://proceedings.iclr.cc/paper_files/paper/2025/file/72163d1c3c1726f1c29157d06e9e93c1-Paper-Conference.pdf}
}

@inproceedings{choromanski2021rethinking,
  title     = {Rethinking Attention with Performers},
  author    = {Choromanski, Krzysztof and Likhosherstov, Valerii and Dohan, David
               and Song, Xingyou and Gane, Andreea and Sarlos, Tamas
               and Hawkins, Peter and Davis, Jared and Mohiuddin, Afroz
               and Kaiser, Lukasz and Belanger, David and Colwell, Lucy
               and Weller, Adrian},
  booktitle = {International Conference on Learning Representations},
  year      = {2021},
  url       = {https://openreview.net/forum?id=Ua6zuk0WRH}
}

@inproceedings{chen2024dijiang,
  title     = {{D}i{J}iang: Efficient Large Language Models through Compact Kernelization},
  author    = {Chen, Hanting and Liu, Zhicheng and Wang, Xutao and Tian, Yuchuan and Wang, Yunhe},
  booktitle = {Proceedings of the 41st International Conference on Machine Learning},
  pages     = {7103--7117},
  year      = {2024},
  volume    = {235},
  series    = {Proceedings of Machine Learning Research},
  publisher = {PMLR},
  url       = {https://proceedings.mlr.press/v235/chen24ab.html}
}

@article{nguyen2026spectraformer,
  author    = {Nguyen, Duke and Yin, Du and Joshi, Aditya and Salim, Flora},
  title     = {Spectraformer: A Unified Random Feature Framework for Transformer},
  journal   = {ACM Transactions on Intelligent Systems and Technology},
  volume    = {17},
  number    = {3},
  articleno = {50},
  numpages  = {29},
  year      = {2026},
  publisher = {Association for Computing Machinery},
  doi       = {10.1145/3768161},
  url       = {https://doi.org/10.1145/3768161}
}

@inproceedings{nauen2024taylorshift,
  title     = {TaylorShift: Shifting the Complexity of Self-Attention from Squared to Linear (and Back) using Taylor-Softmax},
  author    = {Nauen, Tobias Christian and Palacio, Sebastian and Dengel, Andreas},
  booktitle = {Pattern Recognition},
  pages     = {1--16},
  year      = {2024},
  publisher = {Springer Nature Switzerland},
  address   = {Cham},
  doi       = {10.1007/978-3-031-78172-8_1},
  note      = {Proceedings of the 27th International Conference on Pattern Recognition (ICPR 2024)}
}

@article{massive_activation,
  title   = {Massive Activations in Large Language Models},
  author  = {Sun, Mingjie and Chen, Xinlei and Kolter, J. Zico and Liu, Zhuang},
  journal = {arXiv preprint arXiv:2402.17762},
  year    = {2024},
  url     = {https://arxiv.org/abs/2402.17762}
}

@inproceedings{attention_sink,
  title     = {Efficient Streaming Language Models with Attention Sinks},
  author    = {Xiao, Guangxuan and Tian, Yuandong and Chen, Beidi and Han, Song and Lewis, Mike},
  booktitle = {The Twelfth International Conference on Learning Representations},
  year      = {2024},
  url       = {https://openreview.net/forum?id=NG7sS51zVF}
}

@inproceedings{hu2022lora,
  title     = {{LoRA}: Low-Rank Adaptation of Large Language Models},
  author    = {Hu, Edward J. and Shen, Yelong and Wallis, Phillip
               and Allen-Zhu, Zeyuan and Li, Yuanzhi and Wang, Shean
               and Wang, Lu and Chen, Weizhu},
  booktitle = {International Conference on Learning Representations},
  year      = {2022},
  url       = {https://openreview.net/forum?id=nZeVKeeFYf9}
}

@article{llama2024herd,
  title   = {The Llama 3 Herd of Models},
  author  = {{Llama Team}},
  journal = {arXiv preprint arXiv:2407.21783},
  year    = {2024},
  url     = {https://arxiv.org/abs/2407.21783}
}

@inproceedings{kovaleva2021bert,
  title     = {{BERT} Busters: Outlier Dimensions that Disrupt Transformers},
  author    = {Kovaleva, Olga and Kulshreshtha, Saurabh and Rogers, Anna
               and Rumshisky, Anna},
  booktitle = {Findings of the Association for Computational Linguistics:
               ACL-IJCNLP 2021},
  pages     = {3392--3405},
  year      = {2021},
  publisher = {Association for Computational Linguistics},
  doi       = {10.18653/v1/2021.findings-acl.300},
  url       = {https://aclanthology.org/2021.findings-acl.300/}
}

@inproceedings{dettmers2022llmint8,
  title     = {{LLM.int8()}: 8-bit Matrix Multiplication for Transformers at Scale},
  author    = {Dettmers, Tim and Lewis, Mike and Belkada, Younes
               and Zettlemoyer, Luke},
  booktitle = {Advances in Neural Information Processing Systems},
  volume    = {35},
  pages     = {30318--30332},
  year      = {2022},
  publisher = {Curran Associates, Inc.},
  url       = {https://proceedings.neurips.cc/paper_files/paper/2022/file/c3ba4962c05c49636d4c6206a97e9c8a-Paper-Conference.pdf}
}

@article{arora2024based,
  author = {Arora, Simran and Eyuboglu, Sabri and Zhang, Michael and Timalsina, Aman
            and Alberti, Silas and Zinsley, Dylan and Zou, James and Rudra, Atri
            and R{\'e}, Christopher},
  title = {Simple Linear Attention Language Models Balance the Recall-Throughput Tradeoff},
  journal = {arXiv preprint arXiv:2402.18668},
  year = {2024},
  url = {https://arxiv.org/abs/2402.18668}
}

@article{aksenov2024rebased,
  author = {Aksenov, Yaroslav and Balagansky, Nikita and {Lo Cicero Vaina}, Sofia Maria
            and Shaposhnikov, Boris and Gorbatovski, Alexey and Gavrilov, Daniil},
  title = {Linear Transformers with Learnable Kernel Functions are Better In-Context Models},
  journal = {arXiv preprint arXiv:2402.10644},
  year = {2024},
  url = {https://arxiv.org/abs/2402.10644}
}

@article{chowdhury2022learning,
  title={Learning the Transformer Kernel},
  author={Chowdhury, Sankalan Pal and Solomou, Adamos and Dubey, Kumar Avinava and Sachan, Mrinmaya},
  journal={Transactions on Machine Learning Research},
  year={2022},
  url={https://openreview.net/forum?id=tLIBAEYjcv}
}

@inproceedings{lan2025liger,
  title={Liger: Linearizing Large Language Models to Gated Recurrent Structures},
  author={Lan, Disen and Sun, Weigao and Hu, Jiaxi and Du, Jusen and Cheng, Yu},
  booktitle={Proceedings of the 42nd International Conference on Machine Learning},
  pages={32452--32466},
  volume={267},
  series={Proceedings of Machine Learning Research},
  publisher={PMLR},
  year={2025},
  url={https://proceedings.mlr.press/v267/lan25b.html}
}

@article{suzuki2006pvclust,
  title   = {Pvclust: An R Package for Assessing the Uncertainty in Hierarchical Clustering},
  author  = {Suzuki, Ryota and Shimodaira, Hidetoshi},
  journal = {Bioinformatics},
  volume  = {22},
  number  = {12},
  pages   = {1540--1542},
  year    = {2006},
  doi     = {10.1093/bioinformatics/btl117}
}

@article{hennig2007clusterwise,
  title   = {Cluster-wise Assessment of Cluster Stability},
  author  = {Hennig, Christian},
  journal = {Computational Statistics \& Data Analysis},
  volume  = {52},
  number  = {1},
  pages   = {258--271},
  year    = {2007},
  doi     = {10.1016/j.csda.2006.11.025}
}

@inproceedings{peng2021random,
  title     = {Random Feature Attention},
  author    = {Peng, Hao and Pappas, Nikolaos and Yogatama, Dani and Schwartz, Roy and Smith, Noah A. and Kong, Lingpeng},
  booktitle = {International Conference on Learning Representations},
  year      = {2021},
  url       = {https://openreview.net/forum?id=QtTKTdVrFBB}
}

@inproceedings{kasai2021finetuning,
  title     = {Finetuning Pretrained Transformers into {RNN}s},
  author    = {Kasai, Jungo and Peng, Hao and Zhang, Yizhe and Yogatama, Dani
               and Ilharco, Gabriel and Pappas, Nikolaos and Mao, Yi
               and Chen, Weizhu and Smith, Noah A.},
  booktitle = {Proceedings of the 2021 Conference on Empirical Methods in Natural Language Processing},
  pages     = {10630--10643},
  year      = {2021},
  publisher = {Association for Computational Linguistics},
  doi       = {10.18653/v1/2021.emnlp-main.830},
  url       = {https://aclanthology.org/2021.emnlp-main.830/}
}

@inproceedings{kacham2024polysketchformer,
  title     = {{PolySketchFormer}: Fast Transformers via Sketching Polynomial Kernels},
  author    = {Kacham, Praneeth and Mirrokni, Vahab and Zhong, Peilin},
  booktitle = {Proceedings of the 41st International Conference on Machine Learning},
  pages     = {22748--22770},
  year      = {2024},
  volume    = {235},
  series    = {Proceedings of Machine Learning Research},
  publisher = {PMLR},
  url       = {https://proceedings.mlr.press/v235/kacham24a.html}
}

@inproceedings{an2025systematic,
  title     = {Systematic Outliers in Large Language Models},
  author    = {An, Yongqi and Zhao, Xu and Yu, Tao and Tang, Ming and Wang, Jinqiao},
  booktitle = {International Conference on Learning Representations},
  year      = {2025},
  url       = {https://openreview.net/forum?id=rLX7Vyyzus}
}
\bibliographystyle{plainnat}

\clearpage
\appendix

\section{Additional Analysis of Query--Key Heterogeneity}
\label{app:qk_heterogeneity}

This section provides additional evidence for the dimension-wise
heterogeneity discussed in Section~\ref{sec:qk_contribution}. We first
examine the same phenomenon in GPT-2 Medium and under isotropic Gaussian
controls, and then evaluate the stability of the identified
high-contribution structure across inputs and sequence lengths.

\subsection{Cross-Model and Control Evidence}
\label{app:cross_model_control}

\begin{figure}[t]
    \centering
    \begin{subfigure}[t]{0.32\linewidth}
        \centering
        \includegraphics[width=\linewidth]
        {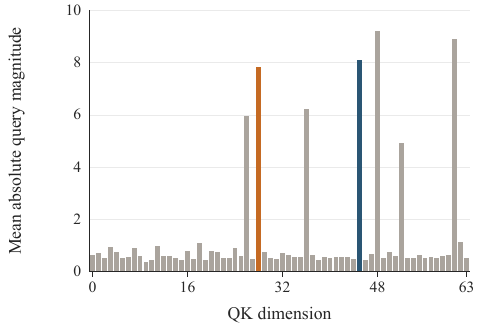}
        \caption{Query magnitude}
    \end{subfigure}
    \hfill
    \begin{subfigure}[t]{0.32\linewidth}
        \centering
        \includegraphics[width=\linewidth]
        {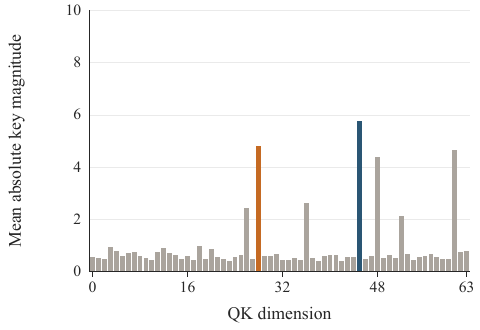}
        \caption{Key magnitude}
    \end{subfigure}
    \hfill
    \begin{subfigure}[t]{0.32\linewidth}
        \centering
        \includegraphics[width=\linewidth]
        {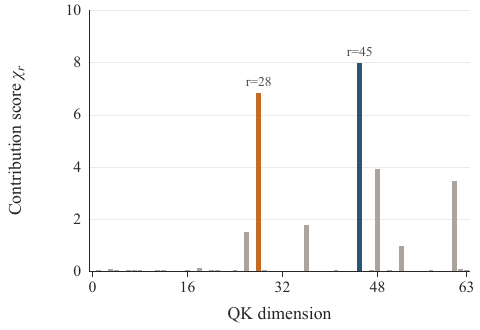}
        \caption{Dimension-wise contribution}
    \end{subfigure}

    \caption{
    \textbf{Dimension-wise heterogeneity in a representative GPT-2
    Medium attention head.}
    GPT-2 exhibits the same qualitative relationship between
    nonuniform query/key representations and concentrated query--key
    contributions as observed in Qwen2.5-0.5B, although the dominant
    contribution is distributed across a broader set of dimensions.
    }
    \label{fig:appendix_gpt2_qk}
\end{figure}

\paragraph{GPT-2 Medium}
We repeat the dimension-wise analysis on GPT-2 Medium to examine
whether the heterogeneity observed in Qwen2.5-0.5B is model-specific.
As shown in Figure~\ref{fig:appendix_gpt2_qk}, GPT-2 also exhibits
nonuniform query/key magnitudes and a strongly concentrated
contribution profile, although the dominant contribution is distributed across a broader set of dimensions.

\paragraph{Gaussian controls}
The Gaussian-control column in
Table~\ref{tab:error_concentration} uses a synthetic counterpart of the
pretrained representations. Specifically, we replace the pretrained
post-RoPE query and key representations with i.i.d.\ standard Gaussian
samples drawn from $\mathcal N(0,1)$, while keeping the remaining
experimental protocol unchanged. We compute the same contribution
statistics, apply the same grouping procedure to identify the
corresponding high-contribution dimensions, and use the same
approximation, intervention, and aggregation procedures as in the
pretrained experiment. Thus, the Gaussian control changes only the
distribution of the query and key representations.

The statistical significance reported in Section~\ref{sec:contribution_error}
is based on a separate matched-control randomization test on the original
pretrained representations, described in Appendix~A.2; it is not computed
from the Gaussian controls.

\begin{figure}[t]
    \centering
    \includegraphics[width=0.48\linewidth]
    {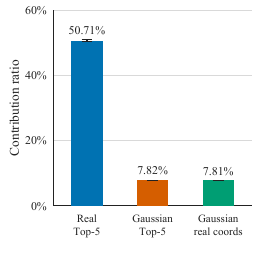}
    \caption{
    \textbf{Top-5 contribution concentration.}
    The pretrained Qwen2.5-0.5B representation exhibits substantially
    stronger concentration than either isotropic Gaussian control.
    }
    \label{fig:appendix_top5_concentration}
\end{figure}

As an additional diagnostic, for the representative Qwen2.5-0.5B
attention head analyzed in Figure~\ref{fig:appendix_top5_concentration},
we compare the top-5 contribution concentration of the pretrained
representations with isotropic Gaussian controls of the same
dimensionality. The Gaussian controls use the same head dimension and
the same contribution statistic $\chi_r$, but replace the pretrained
query--key representations with i.i.d.\ standard Gaussian samples drawn
from $\mathcal N(0,1)$.

The Gaussian controls indicate that the observed concentration is not
explained solely by dimensionality or by selecting particular
coordinate indices.

\subsection{High-Contribution-Dimension Intervention}
\label{app:subset_intervention}

We further specify the intervention used in
Table~\ref{tab:error_concentration}. For each evaluated query head $h$,
let $\mathcal G_{h,1}$ denote the bulk group produced by the fixed OAL
grouping procedure, and let $G_h$ denote the total number of groups.
We define the high-contribution subset as
\[
    \mathcal S_h
    =
    \bigcup_{g=2}^{G_h}\mathcal G_{h,g},
\]
that is, the union of all non-bulk groups for that head. The subset is
determined independently for each layer--query-head pair using only the
calibration contribution statistics.

Across the 42 evaluated query heads, the selected subsets contain
147 dimensions in total. Since $d_h=64$, their aggregate fraction is
\[
    \frac{\sum_h |\mathcal S_h|}{42d_h}
    =
    \frac{147}{42\times64}
    \approx 5.47\%,
\]
which gives the approximately $5.5\%$ reported in
Table~\ref{tab:error_concentration}.

Let $\bar{\mathcal S}_h$ denote the complement of $\mathcal S_h$.
For a causal query--key pair $(i,j)$, the exact softmax kernel
factorizes as
\begin{equation}
    \kappa_{ij}^{\mathrm{exact}}
    =
    \exp\left(
        \frac{q_i^\top k_j}{\sqrt{d_h}}
    \right)
    =
    \kappa_{ij,\mathcal S_h}^{\mathrm{exact}}
    \kappa_{ij,\bar{\mathcal S}_h}^{\mathrm{exact}},
\end{equation}
where
\begin{equation}
    \kappa_{ij,\bar{\mathcal S}_h}^{\mathrm{exact}}
    =
    \exp\left(
        \frac{
            \sum_{r\notin \mathcal S_h} q_{i,r}k_{j,r}
        }{\sqrt{d_h}}
    \right).
\end{equation}

To measure the error associated with approximating only $\mathcal S_h$, we mask
the query and key coordinates outside $\mathcal S_h$ and apply the corresponding
baseline approximation to the masked vectors, retaining the kernel's
input preprocessing. The approximate factor for the selected dimensions
and the exact factor for their complement are computed separately.
Denoting the resulting
sub-kernel by $\widehat \kappa_{ij,\mathcal S_h}^{(m)}$ for method $m$, we form the
full-head hybrid kernel
\begin{equation}
    \widetilde \kappa_{ij}^{(m,\mathcal S_h)}
    =
    \widehat \kappa_{ij,\mathcal S_h}^{(m)}
    \kappa_{ij,\bar{\mathcal S}_h}^{\mathrm{exact}} .
    \label{eq:subset_hybrid_kernel}
\end{equation}
Thus, dimensions outside $\mathcal S_h$ remain exact at the unnormalized-kernel
level. The hybrid kernel is then causally masked, normalized over the
key positions, and applied to the pretrained values:
\begin{equation}
    \widetilde y_i^{(m,\mathcal S_h)}
    =
    \frac{
        \sum_{j\leq i}
        \widetilde \kappa_{ij}^{(m,\mathcal S_h)} v_j
    }{
        \sum_{j\leq i}
        \widetilde \kappa_{ij}^{(m,\mathcal S_h)}
    } .
    \label{eq:subset_hybrid_output}
\end{equation}
The tilde distinguishes this hybrid output from a full-head approximation.
The same intervention is used for all evaluated approximation methods;
the methods differ only in the construction of
$\widehat \kappa_{ij,\mathcal S_h}^{(m)}$.

\paragraph{Error metric}
The intervention is evaluated at the attention-output level rather than
on kernel values or attention probabilities. Let
$Y^{\mathrm{softmax}}$ denote the exact causal softmax attention output.
For an evaluated input and layer, we define
\begin{align}
    \mathcal E_{\mathcal S}^{(m)}
    &=
    \operatorname{MSE}\!\left(
        \widetilde Y^{(m,\mathcal S_h)},
        Y^{\mathrm{softmax}}
    \right),\\
    \mathcal E_{\mathrm{full}}^{(m)}
    &=
    \operatorname{MSE}\!\left(
        \widehat Y^{(m,\mathrm{full})},
        Y^{\mathrm{softmax}}
    \right),
\end{align}
where the MSE is averaged over the evaluated query heads, query
positions, and output-head dimensions. We report the retained-error
ratio
\begin{equation}
    R_{\mathcal S}^{(m)}
    =
    \frac{
        \mathcal E_{\mathcal S}^{(m)}
    }{
        \mathcal E_{\mathrm{full}}^{(m)}
    } .
    \label{eq:subset_retained_error}
\end{equation}
A larger value therefore indicates that restricting the approximation
to the selected dimensions preserves a larger fraction of the
full-method attention-output error.

\paragraph{Evaluation protocol}
We evaluate Qwen2.5-0.5B on zero-based Layers 4, 12, and 20, using all
query heads in each layer. Evaluation uses 16 disjoint
4096-token WikiText-2 sequences, with 32 causal query positions sampled
per sequence and layer. For stochastic approximation methods, we use
five random seeds, $\{0,1,2,3,4\}$. The selected-subset and full-method
MSEs are first averaged over these seeds before forming the
retained-error ratio. TaylorShift is deterministic and is evaluated
once. The resulting ratios are then macro-averaged across the evaluated
layers and input sequences.

\paragraph{Matched-control randomization test}
To assess whether the retained-error ratio of the high-contribution
subset can be explained by selecting an equally sized subset of
ordinary dimensions, we construct $B=1000$ matched control subsets.
For each control, the same intervention and aggregation procedure is
used to obtain a retained-error ratio
$R_b^{\mathrm{control}}$.
We perform a one-sided Monte Carlo randomization test and report
\begin{equation}
    p
    =
    \frac{
        1+\sum_{b=1}^{B}
        \mathbf{1}
        \left[
            R_b^{\mathrm{control}}
            \geq
            R_{\mathcal S}
        \right]
    }{
        B+1
    } .
    \label{eq:subset_randomization_test}
\end{equation}
The $+1$ correction avoids reporting a zero Monte Carlo $p$-value;
with $B=1000$, the minimum reportable value is
$1/1001\approx0.000999$.

\subsection{Stability of Dimension-Wise Contributions}
\label{app:qk_stability}

\begin{figure}[t]
    \centering
    \includegraphics[width=0.52\linewidth]
    {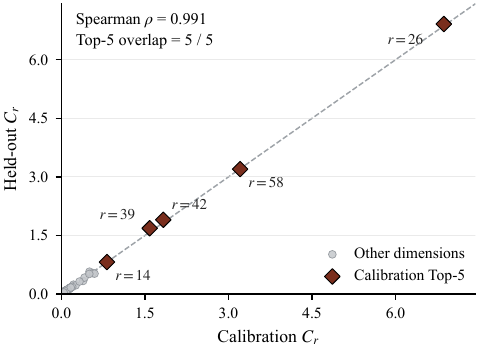}
    \caption{
    \textbf{Stability of dimension-wise contributions across inputs.}
    Contribution scores computed from calibration and held-out inputs
    achieve Spearman correlation $\rho=0.991$, with a top-5 overlap of
    $5/5$.
    }
    \label{fig:input_stability}
\end{figure}

\paragraph{Across inputs}
We first examine whether the contribution profile depends strongly on
the calibration inputs. Figure~\ref{fig:input_stability} compares the
dimension-wise contribution scores computed from calibration and
held-out inputs for the representative Qwen2.5-0.5B attention head.
The two profiles are highly consistent, with Spearman correlation
$\rho=0.991$ and identical top-5 dimensions.

\paragraph{Across sequence lengths}
We further evaluate the same representative head using prefixes of a
fixed sequence at different lengths. Table~\ref{tab:seqlen_stability}
reports the normalized contributions of the five dimensions identified
at sequence length $4096$.

\begin{table}[t]
    \centering
    \normalsize
    \caption{
    \textbf{Stability across sequence lengths.}
    Normalized contribution $\chi_r/\sum_t\chi_t$ (\%) of the five
    highest-contribution dimensions identified at sequence length
    $4096$ for the representative attention head.
    }
    \label{tab:seqlen_stability}
    \begin{tabular}{@{}cccccc@{}}
        \toprule
        Length &
        $r=26$ &
        $r=58$ &
        $r=19$ &
        $r=39$ &
        $r=7$ \\
        \midrule
        128  & \textbf{33.34} & \textbf{13.64} & 2.12 & 3.30 & 3.56 \\
        512  & \textbf{32.96} & \textbf{15.70} & 1.27 & 3.66 & 3.86 \\
        1024 & \textbf{32.19} & \textbf{15.91} & 1.27 & 3.78 & 3.86 \\
        2048 & \textbf{30.93} & \textbf{16.18} & 2.10 & 3.75 & 3.80 \\
        4096 & \textbf{28.75} & \textbf{17.53} & 4.11 & 3.71 & 3.66 \\
        \bottomrule
    \end{tabular}
\end{table}

The two dominant dimensions, $r=26$ and $r=58$, remain the largest
contributors across all evaluated sequence lengths and jointly account
for approximately $46$--$49\%$ of the total contribution. Together
with the held-out-input analysis, these results indicate that the
dominant dimension-wise structure is stable under changes in both input
content and sequence length.

\section{Grouping and Coefficient-Calibration Details}
\label{app:grouping_calibration_details}

\paragraph{Dimension grouping}
For a fixed layer and query head, we compute the dimension-wise
contribution scores $\chi_r$ from the calibration set as defined in
Eq.~\eqref{eq:qk_contribution}. The queries and keys are the
post-RoPE representations used by the attention module, causal
query--key pairs are used throughout, and query heads follow the
model's native GQA mapping to key--value heads. Different prompts are
aggregated independently and then averaged with equal weight.

We sort the $d_h$ dimensions by increasing contribution,
\[
    \chi_{\pi_1}
    \leq
    \chi_{\pi_2}
    \leq
    \cdots
    \leq
    \chi_{\pi_{d_h}},
\]
using a stable ordering, and search for group boundaries only in the
high-contribution tail. Given a tail fraction $\alpha$, let
\[
    m=\lceil \alpha d_h\rceil.
\]
Candidate cuts are restricted to
\[
    k\in
    \left\{
        \max(1,d_h-m),\ldots,d_h-1
    \right\}.
\]
A cut at position $k$ separates
\[
    \mathcal L_k=\{\pi_1,\ldots,\pi_k\},
    \qquad
    \mathcal U_k=\{\pi_{k+1},\ldots,\pi_{d_h}\}.
\]
Thus, the cut immediately preceding the top-$m$ tail is included, while
the endpoint $k=d_h$ is excluded so that both sides of every candidate
cut are nonempty.

For each candidate cut, we first measure the adjacent contribution gap
using
\begin{equation}
    \rho_k
    =
    \frac{
        \chi_{\pi_{k+1}}
    }{
        \max(\chi_{\pi_k},\varepsilon)
    },
    \label{eq:gap_ratio}
\end{equation}
where $\varepsilon$ is a small numerical constant used only to avoid
division by zero. A large $\rho_k$ indicates a sharp increase in
contribution across the candidate boundary.

To test whether this separation is stable with respect to the calibration
inputs, we perform $B$ prompt-level bootstrap replicates. For replicate
$b$, let
\[
    I_b=(i_1^{(b)},\ldots,i_P^{(b)})
\]
denote the length-$P$ sequence of prompt indices sampled independently
with replacement from $\{1,\ldots,P\}$. Repeated indices are retained
with their sampling multiplicities. We then recompute the contribution
scores as
\[
    \chi_r^{(b)}
    =
    \frac{1}{P}
    \sum_{t=1}^{P}
    \chi_{i_t^{(b)},r}.
\]
Importantly, the
dimension ordering $\pi$ obtained from the full calibration set is
kept fixed across bootstrap replicates. For candidate cut $k$, we
measure the separation between the two sides as
\begin{equation}
    R_k^{(b)}
    =
    \frac{
        \min_{r\in\mathcal U_k} \chi_r^{(b)}
    }{
        \max\!\left(
            \max_{r\in\mathcal L_k} \chi_r^{(b)},
            \varepsilon
        \right)
    }.
    \label{eq:bootstrap_gap}
\end{equation}
Its bootstrap support is then
\begin{equation}
    S_k
    =
    \frac{1}{B}
    \sum_{b=1}^{B}
    \mathbf{1}
    \left[
        R_k^{(b)}\geq\tau
    \right],
    \label{eq:bootstrap_support}
\end{equation}
which measures the fraction of bootstrap samples for which the two
sides remain separated by at least a factor of $\tau$.

With minimum bootstrap support $\gamma$, a candidate boundary is retained
as a \emph{natural cut} when
\[
    \rho_k\geq\tau
    \qquad\text{and}\qquad
    S_k\geq\gamma.
\]
All natural cuts satisfying these conditions are retained; we do not
apply an additional minimum group size, boundary-spacing constraint, or
maximum group-count cap.

To prevent the grouped parameterization from collapsing to the global
single-group case, we require at least $G_{\min}$ groups per head. If
the natural cuts yield fewer than $G_{\min}$ groups, unused candidate
cuts are added in descending lexicographic order of
$(S_k,\rho_k)$ until the minimum group count is reached. Bootstrap
support is therefore prioritized over the full-sample gap magnitude;
exact ties retain the original candidate-cut order. These
\emph{fallback cuts} are restricted to the same high-contribution tail
but need not satisfy the natural-cut thresholds.

Let the final retained cuts be
$k_1<\cdots<k_{G-1}$, with $k_0=0$ and $k_G=d_h$.
The resulting groups are
\begin{equation}
    \mathcal G_g
    =
    \{
        \pi_{k_{g-1}+1},
        \ldots,
        \pi_{k_g}
    \},
    \qquad
    g=1,\ldots,G.
    \label{eq:dimension_groups}
\end{equation}
Thus, the first group typically forms a large low-contribution shared
bulk, while subsequent groups refine the high-contribution tail.
Grouping is performed independently for every layer and query head,
and the resulting assignments are mapped back to the original
dimension indices. The groups are determined before kernel-coefficient
fitting and remain fixed during subsequent calibration, model
adaptation, and evaluation.

\paragraph{Grouping settings}
Both backbones use the same fixed set of 40 short calibration prompts
and $B=1000$ prompt-level bootstrap replicates. For the formal
experiments, Qwen2.5-0.5B uses
$(\alpha,\gamma)=(0.25,0.9)$, while Llama-3.2-1B uses
$(\alpha,\gamma)=(0.50,0.5)$. Both use
$\tau=1.5$ and $G_{\min}=2$, with no explicit maximum group-count
constraint.

\paragraph{Coefficient calibration}
For a fixed layer and query head, let $\mathcal C$ be the coefficient-fitting
sequences and $\widehat y_i(L;x)$ the normalized causal OAL output for
$x\in\mathcal C$. With groups and pretrained queries, keys, and values fixed,
we fit the lower-triangular $L$ to the corresponding softmax outputs:
\begin{equation}
    L^\star
    =
    \arg\min_L
    \frac{1}{|\mathcal{C}|}
    \sum_{x\in\mathcal{C}}
    \sum_i
    \left\|
        \widehat{y}_i(L;x)-y_i^{\mathrm{softmax}}(x)
    \right\|_2^2.
    \label{eq:oal_calibration}
\end{equation}
Matching the normalized output directly accounts for both the kernel
weights and their normalization denominator. The calibrated factors
$L^\star$ initialize subsequent model-level adaptation, during which
the dimension groups remain fixed while $L$ is jointly optimized with
the LoRA parameters under the causal language-modeling objective.

\paragraph{Coefficient-calibration settings}
Coefficient fitting uses WikiText-2 raw v1 training text, concatenated
and divided into non-overlapping sequences of length $4096$. We use
blocks $16$--$23$ as eight training sequences and blocks $24$--$25$ as
two validation sequences. This coefficient-fitting data is separate
from the short-prompt calibration data used to determine the dimension
groups.

We optimize the OAL factors with Adam at learning rate $0.02$, without
a learning-rate scheduler, for up to $1000$ updates per initialization.
Parameters are optimized in FP32 with gradient norm clipped at $10$.
Each update samples one training sequence and $32$ causal query
positions. Validation is performed every $100$ updates, with early
stopping after $400$ updates without improvement.

We use three initialization candidates for OAL: an embedding of the
learned global quadratic into the grouped parameterization, a
second-order Taylor initialization, and a Gaussian-perturbed
learned-global initialization. Within each initialization, the
lowest-validation-MSE checkpoint is retained independently for each
query head; the final coefficients are then selected per head across
the three initialization candidates.

Layers are calibrated separately, with independent coefficients for
each query head. The learned global quadratic uses the same training
and validation tensors, output-matching objective, optimizer, learning
rate, query-position sampling protocol, and checkpoint selection rule
as OAL. Both methods use three initialization candidates with up to
$1000$ updates per initialization; they differ only in their kernel
parameterization and corresponding initializations.

Frozen-attention evaluation uses blocks $0$--$15$ of the same token
stream, giving 16 additional sequences of length $4096$. These blocks
are disjoint from both coefficient training and validation and are not
used for coefficient optimization or checkpoint selection.

\paragraph{Group-budget diagnostic}
Figure~\ref{fig:grouping_overview}(b) uses the same representative
Qwen2.5-0.5B query head (zero-based Layer 4, Head 0) for all group
budgets. Dimensions are sorted by the same calibration contribution
scores $\chi_r$, and for a prescribed budget $G$ we select the $G-1$
largest adjacent gaps in the sorted profile as group boundaries.
Thus, increasing $G$ only further subdivides the same contribution
ordering.

All budgets use the same coefficient-fitting and evaluation data:
eight 4096-token WikiText-2 blocks for training, two for validation,
and 16 disjoint blocks for held-out evaluation. Coefficients are fitted
with the same Adam-based optimization sweep and selected by validation
attention-output MSE; held-out data are never used for model selection.
The plotted error is pooled held-out attention-output MSE over all
tokens and output dimensions. Therefore, the points in
Figure~\ref{fig:grouping_overview}(b) are directly comparable across group
budgets, with the group partition and corresponding parameter count as
the only systematic changes.

\section{Frozen-Attention Error Metrics}
\label{app:frozen_metrics}

For each held-out sequence, let $Y_b$ denote the exact softmax
attention output and $\widehat{Y}_b$ the corresponding approximation,
with all evaluated layers and heads collected into the same tensor.
We report

\begin{equation}
\mathrm{MSE}
=
\frac{1}{B}
\sum_{b=1}^{B}
\frac{
\left\|\widehat{Y}_b-Y_b\right\|_F^2
}{
n_b
},
\end{equation}

and

\begin{equation}
\mathrm{RelMSE}
=
\frac{1}{B}
\sum_{b=1}^{B}
\frac{
\left\|\widehat{Y}_b-Y_b\right\|_F^2
}{
\left\|Y_b\right\|_F^2
},
\end{equation}

where $B$ is the number of held-out sequences and $n_b$ is the number
of scalar output elements in sequence $b$. Thus, RelMSE normalizes the
squared approximation error by the energy of the exact softmax output
for each sequence before averaging across sequences.

For the metrics before $W_o$, $Y_b$ and $\widehat{Y}_b$ contain the per-head
attention outputs before the output projection. For the
post-$W_o$ metrics, the heads are first concatenated in their original
order and passed through the frozen output projection $W_o$; the same
MSE and RelMSE definitions are then applied to the projected outputs.
The post-$W_o$ RelMSE is therefore normalized by the energy of the
exact softmax output after $W_o$, rather than by the pre-projection
attention output.

\section{Global Reference and HD Realization}
\label{app:oal_realization}

\paragraph{Positive global reference}
The learned global reference is a quadratic in the total score
$s=q^\top k/\sqrt{d_h}$. A symmetric positive semidefinite matrix
$M_{\mathrm{global}}=[m_{uv}]_{u,v=0}^{1}\in\mathbb R^{2\times2}$
parameterizes it as
\begin{equation}
    \begin{aligned}
    \kappa_{\mathrm{global}}(q,k)
    &=\epsilon+
       \begin{bmatrix}1&s\end{bmatrix}
      M_{\mathrm{global}}
      \begin{bmatrix}1&s\end{bmatrix}^{\!\top}\\
    &=(\epsilon+m_{00})+2m_{01}s+m_{11}s^2,
    \qquad M_{\mathrm{global}}\succeq0.
    \end{aligned}
    \label{eq:global_positive_quadratic}
\end{equation}
Thus this is the global polynomial $a+bs+cs^2$ with
$a=\epsilon+m_{00}$, $b=2m_{01}$, and $c=m_{11}$; the matrix constraint
and $\epsilon>0$ ensure strictly positive weights.
It is contained in OAL: setting $M_{00}=m_{00}$, $M_{0g}=M_{g0}=m_{01}$,
and $M_{gh}=m_{11}$ for all groups recovers the same polynomial because
$s=\sum_gs_g$. This grouped matrix is positive semidefinite, since its
quadratic form on any vector $t$ equals the quadratic form of
$M_{\mathrm{global}}$ on $[t_0,\sum_gt_g]^\top$, and hence admits
$M=LL^\top$.

\paragraph{Evaluation equivalence}
For a single query, $(q_i^\top k_j)^2=q_i^\top k_jk_j^\top q_i$ gives
\[
    \sum_j(q_i^\top k_j)^2V_{j,\ell}
    =q_i^\top\left(\sum_jV_{j,\ell}k_jk_j^\top\right)q_i
    =q_i^\top K^\top D_{\ell}Kq_i.
\]
The last expression is the $i$th diagonal entry of $QK^\top D_{\ell}KQ^\top$,
proving Eq.~\eqref{eq:hd_diagonal_identity}. Multiplication by $D_{\ell}$
amounts to scaling key rows and does not require storing a diagonal matrix.
For the grouped kernel, expanding $M=LL^\top$ gives
\begin{equation}
    \begin{aligned}
    \kappa_{\mathrm{OAL}}(q,k)
    &=\epsilon+M_{00}+2\sum_gM_{0g}s_g+\sum_{g,h}M_{gh}s_gs_h\\
    &=a+\frac{1}{\sqrt{d_h}}\sum_r b_rq_rk_r
      +\frac{1}{d_h}\sum_{r,t}C_{rt}q_rq_tk_rk_t,
    \end{aligned}
\end{equation}
which yields the coefficient mapping in Section~\ref{sec:oal_algorithm}.
Substituting Eq.~\eqref{eq:oal_hd_states} into
Eq.~\eqref{eq:oal_hd_readout} and exchanging sums then gives
\begin{equation}
    \begin{aligned}
    \widetilde u_i
    &=\sum_{j\leq i}\left(a+\frac{1}{\sqrt{d_h}}\sum_r b_rq_{i,r}k_{j,r}
      +\frac{1}{d_h}\sum_{r,t}C_{rt}q_{i,r}q_{i,t}k_{j,r}k_{j,t}\right)u_j\\
    &=\sum_{j\leq i}\kappa_{\mathrm{OAL}}(q_i,k_j)u_j.
    \end{aligned}
\end{equation}
The same substitution establishes the identity for
$\kappa_{\mathrm{dim}}$ with general coefficients. The argument uses
separable query and key factors, not Mercer positive semidefiniteness;
pointwise positivity of OAL's weights does not imply the latter property.

\paragraph{Exact state and parameter counts}
The ordered-pair states in Eq.~\eqref{eq:oal_hd_states} contain
$(1+d_h+d_h^2)(d_v+1)$ entries. Since $H_{2,i,rt}=H_{2,i,tr}$ and
$C$ is symmetric, the packed-symmetric implementation stores only
$(1+d_h+d_h(d_h+1)/2)(d_v+1)$ entries. Diagonal pairs have multiplicity
one and off-diagonal pairs multiplicity two during readout, preserving
all interactions. These counts describe one logical state, not total
workspace or peak memory; GQA can share KV states across query heads.
The lower-triangular factor $L$ has $(G+1)(G+2)/2$ learned parameters
per head.

\section{Runtime and Memory Benchmark}
\label{app:runtime_benchmark}

Figure~\ref{fig:oal_runtime_memory} evaluates forward-only causal
attention on an NVIDIA A100-SXM4-40GB. The batch size is $B=1$, with
$H_q=14$ query heads, $H_{kv}=2$ key/value heads, and
$d_h=d_v=64$. Input tensors use BF16, and sequence lengths range from
$2^{10}$ to $2^{15}$. These are attention-operator measurements, not
end-to-end model latency or per-token autoregressive decoding latency.

OAL ($G=5$) and Learned Global ($G=1$) use the same optimized HD
block-GEMM backend, with 128-token blocks, BF16 Tensor Core contractions,
and FP32 working coefficients. The implementation materializes
intermediate tensors and reuses structural preparation across timed
calls. OAL uses a fixed five-group configuration, matching the largest
group count observed across the heads used in our model-adaptation
experiments, to measure operator scaling. This benchmark does not execute the heterogeneous group
configurations of all calibrated heads in a converted model.

Performer, SpectraFormer, DiJiang, and TaylorShift use causal
adaptations of their official source code. The source repositories and kernel
adaptations are described in Appendix~\ref{app:baseline_adaptations}.
These methods retain their respective causal execution paths, while
OAL and Learned Global share the specialized HD backend. The
implementations differ in optimization level; in particular, the
TaylorShift adaptation receives no additional kernel optimization.

Internal arithmetic also differs across implementations. Performer
and SpectraFormer use FP64 feature computation and prefix
accumulation. DiJiang uses FP32 feature computation and prefix
accumulation with 128-token blocks, explicitly repeating shared keys
and values across query heads. Together with the backend differences
above, these precision choices affect the measured runtime and memory,
so cross-method gaps cannot be attributed solely to the kernel formulations.

Both softmax references use PyTorch scaled dot-product attention
(SDPA). The FlashAttention curve selects the
\texttt{FLASH\_ATTENTION} backend, whereas the PyTorch SDPA curve
selects the \texttt{MATH} backend. Both calls enable grouped-query
attention and causal masking, with zero attention dropout.
Measurements run with gradient recording disabled, after warmup,
and runtime is wall-clock latency with CUDA synchronization.
Runtime markers report the mean across trials.

GPU memory is the total peak allocated tensor memory reported by
\nolinkurl{torch.cuda.max_memory_allocated()}, including resident inputs
and method state. It is neither incremental memory above the resident
allocation nor allocator-reserved memory. This peak includes workspace
and therefore differs from the logical recurrent-state size analyzed in
Section~\ref{sec:oal_complexity}. For both SpectraFormer and math-backend
SDPA, the $N=16384$ case is skipped by a pre-run memory guard because
the estimated next peak exceeds 80\% of the current free GPU memory.
Both sweeps therefore end at $N=8192$; these omissions reflect the
benchmark's memory budget rather than observed out-of-memory failures.

\section{Conversion and LoRA Setup}
\label{app:conversion_setup}

\begin{figure}[t]
    \centering
    \begin{subfigure}[t]{0.48\linewidth}
        \centering
        \includegraphics[width=\linewidth]{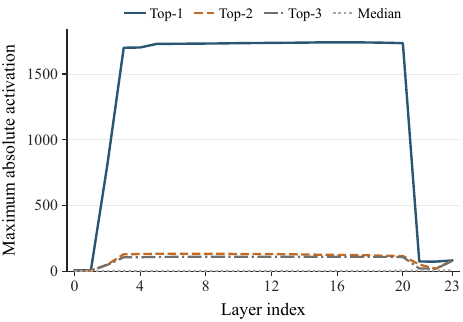}
        \caption{Qwen2.5-0.5B}
        \label{fig:qwen_layer_selection}
    \end{subfigure}
    \hfill
    \begin{subfigure}[t]{0.48\linewidth}
        \centering
        \includegraphics[width=\linewidth]{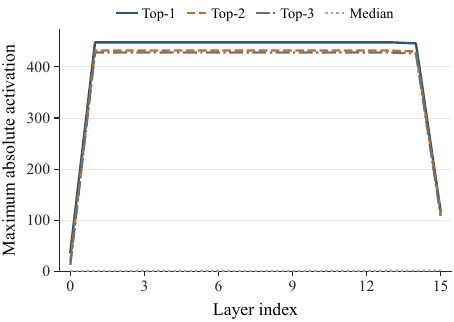}
        \caption{Llama-3.2-1B}
        \label{fig:llama_layer_selection}
    \end{subfigure}
    \caption{\textbf{Layer-wise activation profiles.}
    We measure the output residual stream of each decoder block on the same
    40 short prompts. For each hidden dimension, we record its maximum
    absolute activation over all tokens and prompts; Top-1/2/3 denote the
    three largest dimension-wise maxima, and Median denotes their median
    across dimensions. Layer indices are zero-based.}
    \label{fig:layer_selection}
\end{figure}

\paragraph{Attention replacement}
Figure~\ref{fig:layer_selection} shows sustained large activations in
the intermediate layers of both models, with different concentration
patterns across hidden dimensions. Because pronounced activation
outliers persist through these intermediate layers, we use this region
as the primary target for attention replacement and determine the
replacement ranges accordingly.

For Qwen2.5-0.5B, we replace zero-based layers 3--20 (inclusive),
retaining softmax attention in the first and last three layers. For
Llama-3.2-1B, we replace layers 1--14 (inclusive), retaining softmax
attention in the first and last layers. The same replacement range is
used for all converted methods within each backbone. The softmax
reference retains softmax attention throughout. These are hybrid
models: the retained softmax layers still contribute quadratic prefill
computation, even though the OAL operator is linear in sequence length.

\subsection{Baseline Implementations and Adaptations}
\label{app:baseline_adaptations}

The baseline implementations in Table~\ref{tab:end_to_end_results} are
adapted from the official source code of Performer, SpectraFormer,
DiJiang, and TaylorShift. We integrate their attention kernels into the
same pretrained backbones, retaining the native query, key, value, and
output projections, RoPE, and GQA head mapping. Each kernel uses causal
prefix aggregation and normalization over the available keys. This
setup compares attention approximations under a common conversion
protocol rather than reproducing each method's original full-model
architecture and training recipe.

For the frozen-attention evaluation, trainable kernel parameters are
first calibrated on a small calibration set while the pretrained Q/K/V
states remain fixed. This calibration is separate from the subsequent
LoRA runs, where the trainable kernel parameters are optimized jointly
with LoRA from the start.

\paragraph{Performer}
We adapt the official FAVOR+ implementation\footnote{\url{https://github.com/google-research/google-research/tree/master/performer}}
using positive orthogonal random features with 250 features per query
head. Gaussian orthogonal projections are sampled independently for
each layer and query head and remain fixed throughout adaptation;
only the LoRA parameters are trained. Queries and keys are scaled by
$d_h^{-1/4}$ before feature evaluation. To stabilize exponential
features, query log-features are shifted by their per-token maximum,
while key log-features use a shared maximum over tokens and features
within each head. These common factors cancel in normalized attention.
Feature computation and causal prefix accumulation use FP64, with the
output cast back to the model dtype.

\paragraph{SpectraFormer}
We adapt the OPRF--FastFoodL variant from the official
repository\footnote{\url{https://github.com/dukenguyenxyz/spectraformer}},
using 65 positive features for $d_h=64$. SpectraFormer supports multiple
combinations of feature functions and projection constructions. We select
OPRF--FastFoodL as a lightweight learnable representative: its structured
projection requires only $3d_h$ trainable kernel parameters per layer.
Each replacement layer has
three trainable FastFood vectors $B,G,S\in\mathbb{R}^{64}$ shared across
its query heads, with a fixed permutation. This gives 192 kernel
parameters per layer, or 3,456 for Qwen2.5-0.5B and 2,688 for
Llama-3.2-1B over their respective replacement layers, excluding LoRA.
The vectors remain trainable but are initialized from the standard
FastFood distributions used in the upstream fixed branch. This avoids
the nearly token-independent features observed with the small-normal
initialization of the learnable branch. We retain the upstream feature
scaling and per-token row-maximum
stabilization followed by an additive $10^{-4}$ feature offset.
The attention aggregation is adapted to causal prefixes, using FP64
feature and accumulation arithmetic and FP32 trainable parameters.

\paragraph{DiJiang}
We adapt the DCT projection and feature-wise softmax construction from
the official repository\footnote{\url{https://github.com/YuchuanTian/DiJiang}}.
The fixed projection combines an orthonormal DCT with diagonal inverse
standard-normal samples. After scaling queries and keys by
$d_h^{-1/4}$, we apply the projection and a softmax over the 64 feature
coordinates. We add positive feature weights $D=\exp(\theta)$ to both
query and key features, with a separate 64-dimensional $\theta$ for
each layer and query head, initialized to zero and trained jointly
with LoRA. We normalize the causal weighted-value sum by the
corresponding sum of kernel weights. The released model's positional
decay factors, GroupNorm, and output gate are omitted to retain the
pretrained attention interface. Features and prefix states use FP32.
Thus, the DiJiang row reports this normalized causal adaptation of the
released feature map.

\paragraph{TaylorShift}
We adapt the official implementation\footnote{\url{https://github.com/tobna/TaylorShift}}
by retaining its query/key $\ell_2$ normalization, fixed polynomial
$1+s+\tfrac12s^2$, and learnable per-head temperature. Here,
$s=\tau\,\widehat q^{\top}\widehat k$, where $\widehat q$ and
$\widehat k$ are unit-normalized vectors. Each layer and query head has
its own temperature $\tau$, initialized to one and trained jointly
with LoRA. We replace noncausal aggregation with causal prefix
evaluation. We disable the optional output-amplitude normalization
($\texttt{output\_normalized=False}$). Its full-sequence factor
$\sqrt{N/d_h}$ would make a fixed prefix's output depend on the total
input length, while replacing $N$ with the current prefix length would
introduce an additional position-dependent scaling rule. We therefore
use no extra output scaling and retain the usual attention
numerator/denominator normalization.

\paragraph{Learned global quadratic}
This controlled reference uses the positive global parameterization
in Appendix~\ref{app:oal_realization}. It shares OAL's output-calibration
protocol, and its calibrated kernel parameters are jointly optimized
with LoRA. It isolates the effect of global versus contribution-based
coefficient sharing within the quadratic family.

\paragraph{Scope of the comparison}
Our comparison focuses on fixed feature maps and kernels with a small
number of trainable parameters, excluding the shared LoRA adapters.
Hedgehog instead learns dense MLP feature maps
\citep{zhang2024hedgehog}. An affine map from $d_h$ input coordinates
to $m$ output coordinates contains $d_hm+m$ parameters; for
$d_h=m=64$, this is 4,160 parameters per map, before accounting for
replication across heads and layers. This differs from the fixed
projections, trainable vectors, and scalar or grouped coefficients
used in our evaluated kernels. Its original attention-weight
distillation objective also compares query--key pairs, with quadratic
scaling in sequence length when evaluated over all causal pairs;
this training cost does not imply quadratic inference complexity.
We do not evaluate Hedgehog under the present protocol, and our
results therefore do not establish OAL's relative performance against
MLP-based feature maps. Extending the comparison would require
specifying their parameter budget and adaptation procedure, including
whether to retain attention-weight distillation or use output matching.

\subsection{LoRA Adaptation}
For all methods, LoRA is applied to the query, key, value, and output
projections in every layer, including layers that retain softmax
attention. We use rank $8$, scaling parameter $\alpha=16$, and dropout
$0.05$, keeping the pretrained backbone weights frozen. Each method is
adapted for 512 optimizer updates on WikiText-103 with sequence length
4096, batch size 1, and no gradient accumulation, corresponding to
2,097,152 input tokens. All runs use a single training seed of 42 and
the same training-block order within each backbone. OAL starts from
training-side calibrated groups and kernel coefficients; its
coefficients are optimized jointly with LoRA while the groups remain fixed.

We use AdamW for LoRA and trainable kernel parameters, with learning
rate $5\times10^{-4}$, weight decay $0.01$, $(\beta_1,\beta_2)=(0.9,0.999)$,
and optimizer $\epsilon=10^{-8}$. The learning rate follows a linear
warmup over the first $5\%$ of updates, followed by cosine decay.
The gradient norm is clipped to $1.0$. Every 128 updates, we compute
next-token NLL on the same fixed 32 WikiText validation blocks and
select the checkpoint with the lowest validation NLL. Final language-modeling
and downstream evaluations use this selected checkpoint; test results
are not used for checkpoint selection.

\subsection{Evaluation Protocol}
\label{app:evaluation_protocol}

Within each backbone, all methods use the same evaluation protocol and
tokenizer, without chat templates or added BOS/EOS tokens.

\paragraph{Perplexity}
We evaluate \texttt{Salesforce/wikitext} (\texttt{wikitext-103-raw-v1},
test split). Text entries are joined in source order with two newlines,
tokenized, and divided into independent, non-overlapping 4096-token
blocks, discarding the incomplete final block. PPL is the exponential
of the mean negative log-likelihood over all scored tokens, with
4095 next-token targets per block.

\paragraph{Downstream tasks}
We report zero-shot accuracy on PIQA validation
(\texttt{lighteval/piqa}, 1,838 examples), HellaSwag validation
(\texttt{Rowan/hellaswag}, 10,042 examples), and ARC-Easy test
(\texttt{allenai/ai2\_arc}, 2,376 examples).
PIQA and ARC-E use \texttt{Question: \{question\}}, followed by a newline
and \texttt{Answer:}. HellaSwag uses the activity label and context,
with the second context segment capitalized, title markers replaced
by periods, and bracketed artifacts removed. Candidates are prefixed
with a space and ranked by their conditional log-likelihood divided
by candidate token count for PIQA/HellaSwag or option-text character
count for ARC-E (excluding the added space). Only candidate tokens
are scored.

\end{document}